\documentclass[lettersize,journal]{IEEEtran}
\usepackage{amsmath,amsfonts}
\usepackage{algorithm}
\usepackage{algorithmicx}
\usepackage{algpseudocode}
\usepackage{array}
\usepackage[caption=false,font=normalsize,labelfont=sf,textfont=sf]{subfig}
\usepackage{textcomp}
\usepackage{stfloats}
\usepackage{url}
\usepackage{verbatim}
\usepackage{graphicx}
\usepackage{cite}
\usepackage{hyperref}
\usepackage{soul}
\soulregister{\ref}7
\soulregister{\eqref}7
\soulregister{\cite}7
\usepackage{color}

\usepackage{cleveref}
\usepackage{bm}
\usepackage{multirow}

\usepackage{url}

\begin{document}

\title{Sampling-Based Hierarchical Trajectory Planning for Formation Flight}

\author{Qingzhao Liu, Bailing Tian, Xuewei Zhang, Junjie Lu, Zhiyu Li*
\thanks{This work
is supported by the National Natural Science Foundation of China under Grants 62273249,
62203415, 62022060, and 62073234 (Corresponding author: Zhiyu Li).

The authors are with the School of Electric and Information Engineer-ing, Tianjin University, Tianjin 300072, China (email: qz\_liu@tju.edu.cn; bailing\_tian@tju.edu.cn; zhangxuewei@tju.edu.cn; lqzx1998@tju.edu.cn; lizhiyu@tju.edu.cn). }

}

\markboth{}%
{Shell \MakeLowercase{\textit{et al.}}: A Sample Article Using IEEEtran.cls for IEEE Journals}


\maketitle

\begin{abstract}
Formation flight of unmanned aerial vehicles (UAVs) poses significant challenges in terms of safety and formation keeping, particularly in cluttered environments. However, existing methods often struggle to simultaneously satisfy these two critical requirements. To address this issue, this paper proposes a sampling-based trajectory planning method with a hierarchical structure for formation flight in dense obstacle environments. To ensure reliable local sensing information sharing among UAVs, each UAV generates a safe flight corridor (SFC), which is transmitted to the leader UAV. Subsequently, a sampling-based formation guidance path generation method is designed as the front-end strategy, steering the formation to fly in the desired shape safely with the formation connectivity provided by the SFCs. Furthermore, a model predictive path integral (MPPI)
based distributed trajectory optimization method is developed as the back-end part, which ensures the smoothness, safety and dynamics feasibility of the executable trajectory. To validate the efficiency of the developed algorithm, comprehensive simulation comparisons are conducted. The supplementary simulation video can be seen at \url{https://www.youtube.com/watch?v=xSxbUN0tn1M}.

\end{abstract}


\section{Introduction}

The utilization of multi-UAV systems has gained significant traction in both industrial and research domains, owing to their inherent robustness and efficiency, surpassing that of individual UAVs. Collaborative efforts among UAVs have successfully accomplished numerous tasks, including swarm surveillance \cite{lin2022robust}, \cite{TITS_Surveillance}, rescue missions \cite{gonccalves2022automatic}, goods delivery \cite{dorling2016vehicle}, \cite{bahnemann2017decentralized}, tracking \cite{guo2022collision}, \cite{TITS_Tracking}, environmental exploration \cite{hui2023dppm}. In the aforementioned applications, it is imperative for the multi-UAV system to form a desired configuration and maintain coordinated movement within that formation. However, in clustered environments, this formation can be disrupted by obstacles, hindering the overall flexibility of the formation. Consequently, formation keeping and trajectory planning modules face a significant challenge in addressing this issue.

\begin{figure}[!t]
\centering
\includegraphics[width=3.2in]{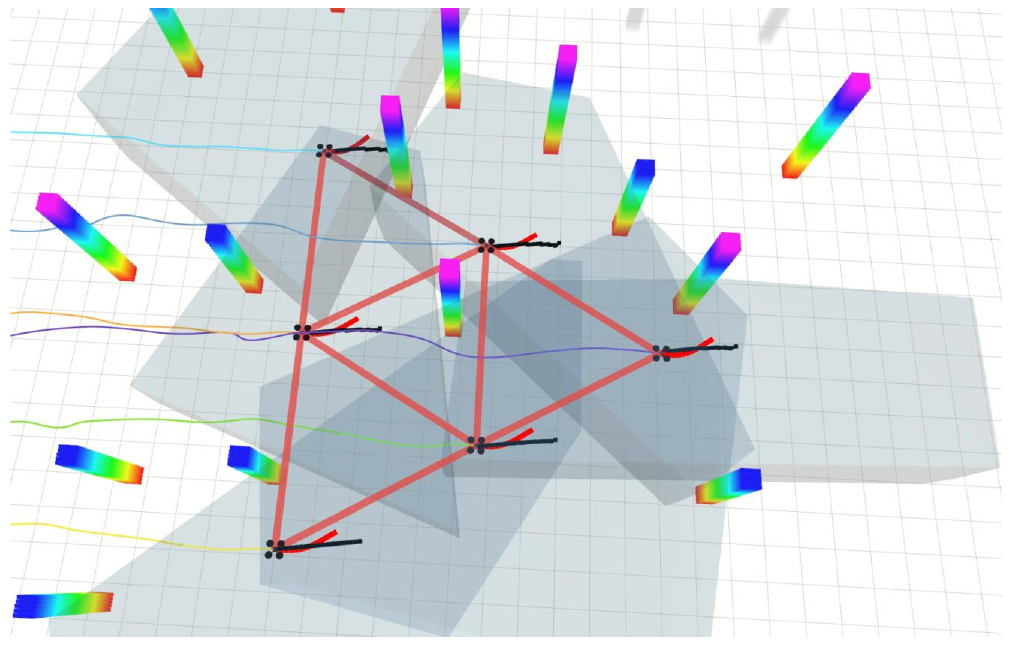}
\caption{A snapshot during the process of formation flight. The light grey polytope is the generated SFC by each UAV, the black lines and light red lines are the formation guidance path generated in Section \ref{formation path} and the trajectory optimized in Section \ref{MPPI} respectively.}
\label{detail_snapshot}
\end{figure}

The majority of existing methods for formation keeping in multi-UAV systems employ control techniques such as position-based control \cite{ren2007distributed}, displacement-based control \cite{wang2020pattern}, and distance-based control \cite{kang2014distance}. However, these methods typically result in rigid formations that lack scalability and are not adaptable to obstacle-rich environments. To address this limitation, a complex Laplacian-based control method with a leader-follower framework has been proposed \cite{han2015formation}. This approach utilizes graph theory \cite{mesbahi2002dynamic} and leverages the complex Laplacian theorem, where accurate formation information is only available to the selected leader, while the other UAVs receive shape information only. For achieving steady-state collective motion, the weight-modification mechanism has been introduced to the complex Laplacian-based formation control \cite{de2021distributed}. Additionally, bearing-based formation control has gained attention due to advancements in sensor technology, allowing for more precise bearing information \cite{wang2022bearing}. However, it is important to note that bearing-based control methods do not address orientation tracking.
To handle more general transformations such as translation, scaling, shearing, or their compositions, the affine formation control approach based on the stress matrix, a type of more general Laplacian matrix, has been proposed \cite{zhao2018affine}. Subsequently, the concept of modularity has been introduced in \cite{Affine_Control_with_odular}, eliminating the need for a designated leader. In this approach, each module defines a cost based on least-squares affine alignment between the positions of the robots in the current and nominal configurations. Although these methods effectively achieve the desired formations with good performance, collision avoidance is not explicitly considered.
In \cite{fathian2019robust}, a distributed control strategy has been devised to autonomously achieve a desired 3D formation in a team of agents, while also accounting for collision avoidance. This robust strategy addresses the potential collisions among the UAVs and ensures safe formation execution.

In addition to traditional control methods, trajectory planning plays a significant role in addressing the formation problem. One approach based on trajectory planning is presented in \cite{turpin2012trajectory}, where a shape-vector is introduced to facilitate the maintenance of the overall formation planning. To enhance practicality, the state estimation module is extended with visual-inertial capabilities, enabling formation execution without relying on an external motion capture system or GPS in \cite{weinstein2018visual}. Besides, in \cite{unknown_payload} the formation trajectories are generated by using the admittance model. Furthermore, a replanning  mechanism is introduced to the formation trajectory planning in \cite{formation_replanning}, which can reconstruct the formation.


The aforementioned methods perform impressively in obstacle-free environments. However, clashes between formation maintenance and obstacle avoidance can occur in clustered environments, which cannot be addressed with those methods. To tackle this issue, Some techniques attempt to co-optimizing formation maintenance with other aspects of trajectory performance. For instance, a virtual-structure based approach for controlling a group of quadrotors carrying out agile interleaved maneuvers is proposed in \cite{zhou2018agile}. A list of potential fields are constructed, including formation preservation, obstacle aversion and collision avoidance fields. The resultant optimal velocity vector is achieved through the sophisticated optimization of these potential fields. Nevertheless, local minima trapping is a noted drawback of the potential field method. 
In contrast, in \cite{morgan2016swarm}, model predictive control (MPC) is employed for the concurrent resolution of optimal assignments and collision-free trajectories. Real-time computations enable instantaneous formation adaptations, allowing for prompt responses to communication or tracking discrepancies.
Taking another route, a differentiable Laplacian-based metric is introduced in \cite{quan2022distributed}, facilitating the quantification of formation keeping performance. However, this approach relies heavily on the weight parameters to coordinate the overlapping effects of formation, obstacle aversion, and internal collisions. 
Despite systematic optimization efforts, resolving conflicts between obstacle avoidance and formation maintenance remains challenging. The paramount importance placed on ensuring the safety of formation flight often leads to compromises in the prioritization of formation keeping, particularly in environments abundant with obstacles.

To avoid the unsatisfactory comprise of formation in the co-optimization, various methods adopt a two-stage framework. In \cite{formation_connectivity}, a valid formation configuration space using an undirected graph is firstly determined, and a graph search methodology is adopted to ascertain a feasible path for the entire system. However, this approach is inherently constrained to environments equipped with a prior map.
In contrast,  in \cite{roy2018multi}  a global-local controller is implemented to ensure obstacle-free formation flight. The global controller is responsible for generating the guiding virtual structure, which serves as a safe region for navigation. Concurrently, the local controller ensures adaptive formation within the dynamic structure that is engendered by the global controller. A similar idea is identifiable in \cite{alonso2015multi}, wherein the formation configuration is primarily determined within the formation polytope. Afterwards, each UAV autonomously navigates toward the designated optimal target independently.
This idea is further expanded in \cite{alonso2016distributed, alonso2017multi}, where the distributive computation is applied to both UAV formation flight and object transportation tasks. in dense obstacle environments, the safe region generated may be overly constrained, which could impede the accommodation of the entire formation at a reduced scale. This limitation could potentially dismantle the formation integrity.
To counteract this issue, the 'split and merge' concept was introduced to enhance formation coordination in \cite{zhu2019distributed}. In this way, the entire formation has the capacity to decompose into smaller sub-formations corresponding to the emergent child regions. Conversely, when the safe region can accommodate the big formation, these sub-formations can seamlessly reassemble into a cohesive full formation.

Although substantial research has been conducted on UAV formation, majority of these studies primarily focus on either formation control \cite{kang2014distance}-\cite{Affine_Control_with_odular} or obstacle and collision avoidance \cite{guo2022collision}, \cite{fathian2019robust}. Besides, the integrated approaches of formation coordination and safety in methods \cite{zhou2018agile}-\cite{zhu2019distributed} cannot guarantee effective formation keeping, especially when the available safe region around the formation is excessively restricted.

\begin{figure}[!t]
\centering
\includegraphics[width=3.4in]{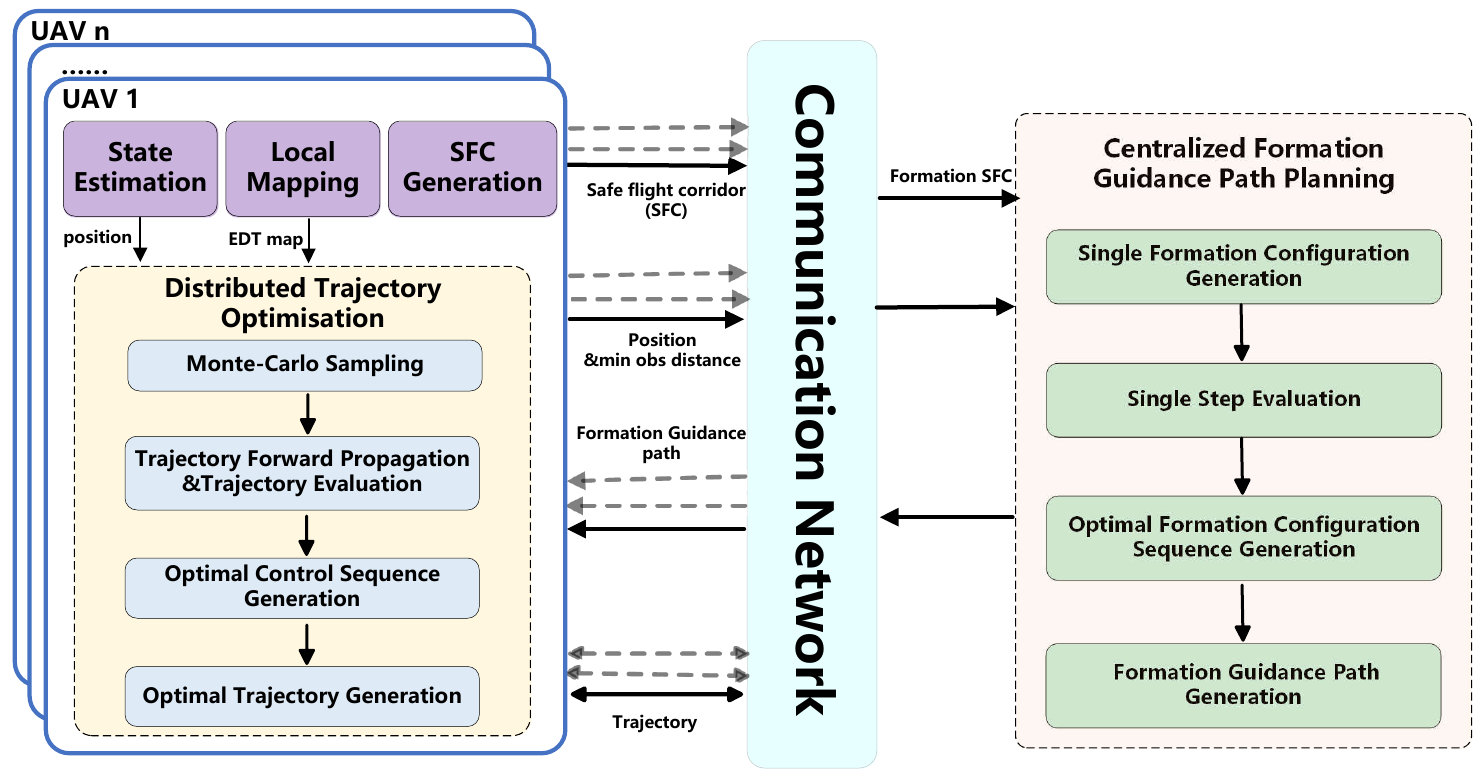}
\caption{Framework of the sampling-based hierarchical trajectory planning method.}
\label{system}
\end{figure}

Motivated by these observations, a hierarchical formation trajectory planning method is proposed in this paper. A snapshot of the formation flight process can be seen in Fig. \ref{detail_snapshot}, where the proposed algorithm is applied. The framework of the proposed method is illustrated in Fig. \ref{system}. Each UAV independently executes the state estimation, local mapping and SFC generation modules. The essential state information, including local position, euclidean distance transform (EDT) and SFC is generated with the local data from sensors. Utilizing the SFC, local position and the minimum distance to the nearest obstacles shared by all the UAVs in the communication network, the formation guidance paths are derived through the centralized formation guidance path planning module running on the leader, which are then broadcast to each UAV from the leader. In contrast to the formation safety limitation with intersection of polytope described in \cite{alonso2016distributed, alonso2017multi}, the union of SFCs from all UAVs is utilized to restrict the whole formation safe region, thereby adapting the clustered environments. Benefiting from the formation connectivity provided by the formation safe region, the safe formation guidance paths ensure the maintenance of formation during the flight process without any collisions with obstacles. Finally, the feasible trajectory is generated locally based on the guidance path, using  the distributed trajectory optimization module for each UAV. The key contributions of this paper can be summarized as follows:

\begin{enumerate}[]
\item A centralized formation guidance path generation method is proposed as the front-end path planning algorithm based on the sampling and receding horizon idea, ensuring flight safety while the formation maintains the desired shape in clustered environments.
\item A distributed trajectory optimization strategy is developed as the back-end trajectory optimization algorithm based on MPPI method, which guarantees the feasibility of the trajectory while following the formation guidance path.
\end{enumerate}


The structure of this article is outlined as follows. Section \ref{pre} offers a comprehensive introduction to fundamental concepts and principles that serve as foundations for comprehending the subsequent algorithm. 
The sampling-based formation guidance path generation and trajectory optimization components are elaborated in Section \ref{formation path} and Section \ref{MPPI}, respectively. The performance of the method is validated through simulations, as presented in Section \ref{validation section}. Finally, the article concludes with a summary and concluding remarks in Section \ref{conclusion}.

\section{PRELIMINARIES} \label{pre}

\subsection{Formation Definition}\label{formation definition}
In this part, some essential definitions are given for subsequent discussions. Considering a formation $\mathcal{F} $ consisting of $N$  UAVs ($N\in \mathbb{Z^+  }$).
The desired formation shape, denoted as $\mathcal{F}_{des}$, is predetermined and represented by a set of $N$ desired individual positions $\bm{p^i_d} = \left[x^i_d, y^i_d, z^i_d\right] \in \mathbb{R}^3$. These positions are defined relative to the formation center $\bm{c} = \left[c_x, c_y, c_z\right] \in \mathbb{R}^3$.
Throughout the flight, the formation center is subject to change as the UAVs progress towards their respective goals, and adjustments in the formation scale may occur. 
Therefore, the formation can also be characterized by the formation configuration $\bm{f_c}$, incorporating a scale factor $s \in \mathbb{R}^+$, the set of desired relative positions $\bm{P_{des}} = \left \{\bm{p^1_d}, \ldots, \bm{p^N_d}\right \}$, and the formation center $\bm{c}$.
The desired relative position set remains constant throughout the entire process, and for simplicity, we exclude the variable $\mathbf{P_{des}}$ from the formation configuration $\bm{f_c}$. Instead, we summarize formations with the same $\bm{P_{des}}$ as similar formations. Therefore, the formation configuration can be represented as $\bm{f_c} = \left \{s, \bm{P_{des}}, \bm{c}\right \}$. Furthermore, the single formation position $\bm{p_f^i}$ for UAV$i$ within a formation can be calculated as follows:
\begin{equation}
\label{single formation position}
\bm{p_f^i} = \bm{c}+  s \bm{p_i^d},i\in \left \{ 1, \cdots, N \right \}
\end{equation}

\subsection{Simplified Description of Obstacle-free Space}\label{SFC section}
The whole 3-D workspace is given by $\mathcal{W} \subseteq \mathbb{R} ^3 $ ,and the obstacle and unknown areas are given by $\mathcal{O}_{obs}$ and $\mathcal{O}_{unk} $ respectively. Therefore, the collision-free workspace is given by $\mathcal{W} _{free}=\mathcal{W}\setminus \mathcal{O}_{obs}\setminus \mathcal{O}_{unk}$. Each UAV $i\in \left \{ 1, \cdots, N \right \} $ in $\mathcal{F} $ can sense its local environment $\mathcal{W}_i$ independently and the local collision-free workspace is denoted as $\mathcal{W}_{free_i}$.

\begin{figure}[!t]
\centering
\subfloat[]{\includegraphics[width=1.5in]{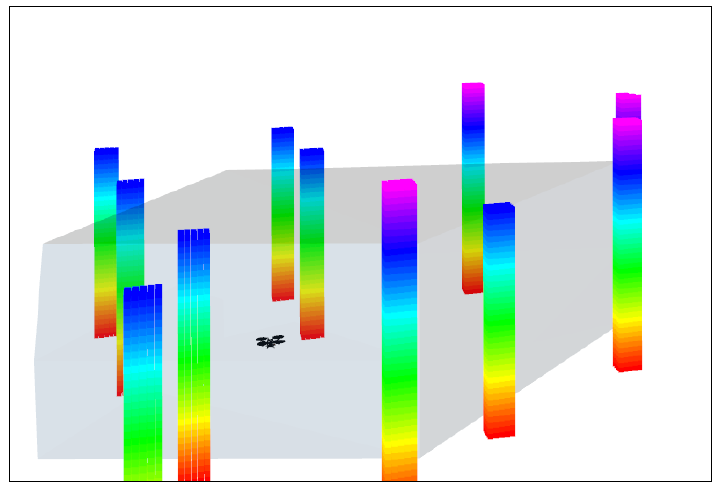}
\label{sfc pic1}}
\hfil
\subfloat[]{\includegraphics[width=1.5in]{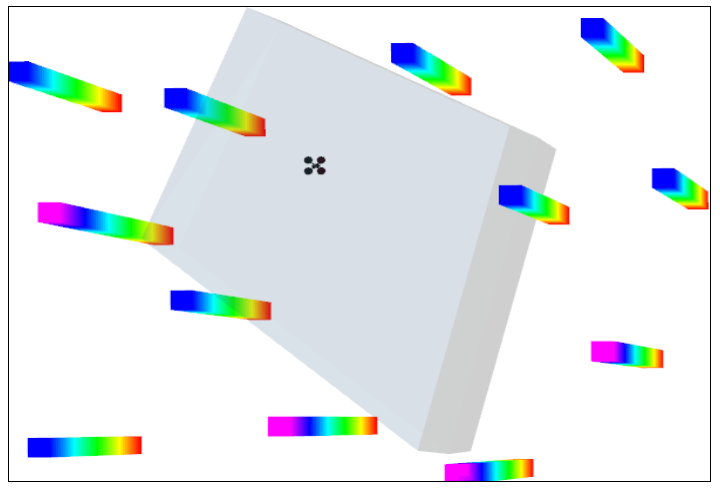}%
\label{sfc pic2}}
\caption{SFC displayed by grey polytope is constructed around the UAV in 3-D environment with some obstacles. Fig. \ref{sfc pic1} and Fig. \ref{sfc pic2} are the main view and the top view of the SFC respectively.}
\label{sfc_illustration}
\end{figure}

The surrounding environment can be recorded with many existing methods of map constructing such as truncated signed distance field map\cite{TSDF}, euclidean signed
distance field map\cite{ESDF} and EDT map\cite{EDT}. However, due to the substantial volume of data contained within the above representations of map, it is impractical to transmit this information in real-world between UAVs.
The 3-D SFC serves as an alternative method to map the environment with limited volume of data, which is a simplified representation of the free space $\mathcal{W}_{free_i}$ surrounding each UAV$i$. 
By setting the current position of each UAV and the local environmental information from sensors as the input, the local SFC is obtained using the fast inflation and recursive intersection algorithm proposed in \cite{wang2024fast}. This algorithm consists of two main steps: ellipsoid inflation and the calculation of the Maximum Volume Inscribed Ellipsoid of the convex polytope, which determines the ellipsoid to be inflated in the subsequent iteration. The SFC, also known as the convex polytope of UAV$i$,  represents the region within which safe flight is ensured, which is illustrated in Fig. \ref{sfc_illustration}. The convex polytope $\mathcal{P}_i$ can be represented as follows: 
\begin{equation}
    \mathcal{P}_i=\left\{\bm{p_x} \in \mathbb{R}^{4} \mid A \bm{p_x} \leq \bm{b}, \text { for } \bm{p_x} \in \mathbb{R}^{n_{l} \times 4}, \bm{b} \in \mathbb{R}^{n_{l}}\right\}
\end{equation}
where $n_l$ denotes the number of faces of $\mathcal{P}_i$, $A$ and $b$ are the hyperplane description parameters of $\mathcal{P}_i$ and $\bm{p_x}$ is the position of point inner $\mathcal{P}_i$.
The only information needed to be shared is the polytope description parameters, whose volume is significantly smaller compared to a local map.

\subsection{Formation Connectivity}\label{formation connectivity}
\begin{figure}[!t]
\centering
\subfloat[]{\includegraphics[width=1.5in]{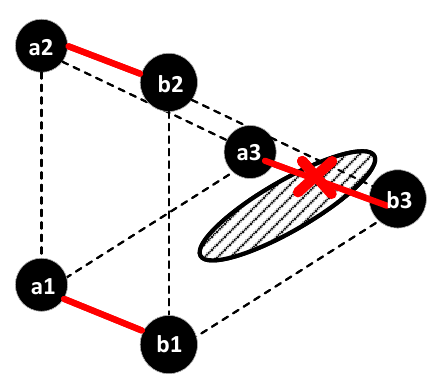}
\label{not connect}}
\hfil
\subfloat[]{\includegraphics[width=1.5in]{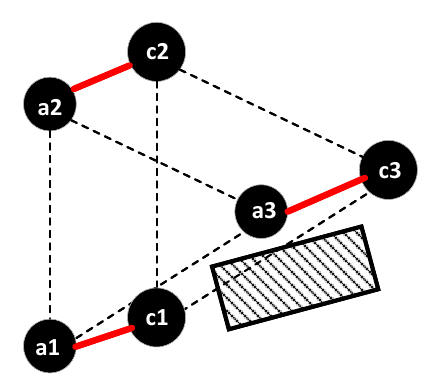}%
\label{connect}}
\caption{The black dot lines connect the inner points in a formation configuration and the red lines are the connecting paths. In Fig. \ref{not connect}, there is no formation connectivity between $\mathbf{f_c^a}$ and $\mathbf{f_c^b}$ due to the intersection of the path between $a3$ and $b3$ with the ellipse obstacle. Conversely, as shown in Fig. \ref{connect}, all paths connecting the correlation points between $\mathbf{f_c^a}$ and $\mathbf{f_c^c}$ are unobstructed, thus indicating that the two formation configurations are connected. }
\label{formation fonnectivity}
\end{figure}

Inspired by the formation configuration connection outlined in definition 2 of \cite{formation_connectivity}, we formulate the formation connectivity based on it, which is illustrated in Fig. \ref{formation fonnectivity}.  Thanks to the safe region confined by the SFC, the formation connectivity can be guaranteed as follows: two formation configurations $f_c^m$ and $f_c^n$ are characterized as connected when they meet the following criteria:
\begin{equation}
\label{formation connectivity equ}
\begin{aligned}
    \bm{p^i_{f_c^m}}  &\in \mathcal{P}_i\\
    \bm{p^i_{f_c^n}}  &\in \mathcal{P}_i
\end{aligned}
\end{equation}
where $\bm{p^i_{f_c^m} }, i\in \{ 1, \cdots, N \}$ represents the formation target for UAV$i$ in $f_c^m$. This constraint establishes the connectivity within the scope of the same local $\mathcal{P}_i$ for all UAVs, ensuring a safe straight path for the complete formation to transition from $f_c^m$ to $f_c^n$. 




\subsection{Trajectory Optimization Problem Formulation}\label{MPPI introduction}
Consider the case where all UAVs adhere to a linear dynamic model in differential flat outputs $\left [ p_x,p_y,p_z,\psi \right ]$, where $p_x$, $p_y$, $p_z$ denote the position in three-dimensional coordinates and $\psi$  represents the yaw. The dynamic model for each UAV $i$ can be written as:
\begin{equation}
\label{dynamics model}
\begin{array}{*{20}{c}}
\bm{\dot x_t^i} = \boldsymbol A\bm{x}_{t}^i + \boldsymbol B\bm{\mu_{t}^i}\\
\boldsymbol A = \left[ {\begin{array}{*{20}{c}}
{{\boldsymbol 0_{6 \times 3}}}& \boldsymbol a\\
{{\boldsymbol 0_{4 \times 3}}}&{{\boldsymbol 0_{4 \times 7}}}
\end{array}} \right],\boldsymbol B = \left[ {\begin{array}{*{20}{c}}
{{\boldsymbol 0_{6 \times 4}}}\\
{{\mathbf{I_{6 \times 6}} }}
\end{array}} \right],\boldsymbol a = {\left[ {\begin{array}{*{20}{c}}
{{\mathbf{I_{6 \times 6}} }}\\
{{\boldsymbol 0_{6 \times 1}}}
\end{array}} \right]^{\text{T}}}
\end{array}
\end{equation}
where $\bm{x}_t^i  = {\left[ {{p_x^i},{p_y^i},{p_z^i},{v_x^i},{v_y^i},{v_z^i},{a_x^i},{a_y^i},{a_z^i},\psi^i } \right]^{\text{T}}}$ is the state vector at time $t$ including position, velocity, acceleration and yaw. Here we let $\bm{\mu_{t}^i} = \bm{u}_{t}^i+\bm{v}_{t}^i$, where $\bm{u}_{t}^i = {\left[ {{{\dot a}_x},{{\dot a}_y},{{\dot a}_z},\dot \psi } \right]^{\text{T}}}$ is the control input and $\bm{v_t^i}\sim \mathcal{N} (0,\Sigma)$ is the stochastic control input following the Gaussian distribution with 0 - mean value and $\Sigma$ - variance. During the stochastic diffusion process, $\bm{v_t^i}$ is applied for generating the forward sampling.

By utilizing the framework of the path integral control\cite{kappen2007introduction}, we can sample a sequence of optimal control input $U_i=\left \{\bm{u}_{t}^i,\dots, \bm{u}_{t+T}^i \right \} $ in a time horizon $T$ to generate a predicted trajectory. In the evaluation of a sampled trajectory, several objectives are considered, including the terminal cost $\phi(\bm{x_}{t+T}, t+T)$, the running cost $\rho(\bm{x}_t, t)$, and the control input cost $\chi(\bm{u_}{t})$. To acquire the optimal trajectory, the above costs are minimized by solving $U_i$. The total cost function of the control problem is defined as follows:
\begin{equation}
\label{value function}
    V(\bm{x}_t^i,t)=\mathbb{E_Q} [\phi(\bm{x}_{t+T}^i,t+T) +\int_{t}^{t+T} (\rho (\bm{x}_\tau ^i,\tau))+\chi(\bm{u}_{\tau}^i)d\tau]
\end{equation}
where $\chi(\bm{u}_{\tau}^i)=\frac{1}{2} (\bm{u}_{\tau}^i)^T\bm{R} \bm{u}_{\tau}^i$, $\bm{R}$ is a positive definite matrix and $\mathbb{E_Q}$ is the mathematical expectation taken by Monte-Carlo sampling.
As a result, the trajectory planning problem can be transformed into a stochastic optimal control problem subject to dynamic model:
\begin{equation}
\label{stochastic optimal control problem}
\begin{array}{cc}
\arg \min\limits_{U_{i}^{*}}  & V(\bm{x}_t^i,t) \\
\text { s.t. } & \bm{\dot x}_t^i = A\bm{x}_{t}^i + B\bm{\mu}_{t}^i
\end{array}
\end{equation}

\section{Sampling-based Formation guidance path Generation Strategy} \label{formation path}

In this section, the sampling-based methodology for generating formation guidance paths is detailed. Firstly, the sampling process used for generating formation configurations is illustrated. Secondly, a comprehensive set of cost functions are introduced, which are utilized to evaluate the obtained formation configurations and formation configuration sequences.  Subsequently, the task assignment optimization is performed to optimize the assignment result which is excluded in the formation configuration sequence. Finally, the formation configuration sequence undergoes a transformation into formation guidance paths using path links.

\subsection{Formation Configuration Sample Process}\label{sample process}
The sampling process is subdivided into two steps to accommodate the dual values inherent in a formation configuration. Initially, a significant quantity of formation center $\bm{c}$ within a hemispherical surface, centered around the local initial formation center $c_0$ are sampled. This approach is implemented through a step propagation mechanism, with a radius specified by $r = (1 + \gamma ) r_{\text{min}}$. where $r_{\text{min}}$ denotes the predefined minimal propagation radius, and $\gamma \in \left [ 0,1 \right ]$. The propagation is strategically directed towards the targe centroid $\bm{{c}_{goal}}$ in the target formation configuration $\bm{f_c^{goal}}$. 
Subsequently, the scale set $S$ is obtained by uniformly generating sacle $s$ values  within a limited range encapsulated by the minimum and maximum scale thresholds $s_{\text{min}}$ and $s_{\text{max}}$. Each of member in $S$ is then paired with a corresponding $\bm{c}$ to form  a plethora of sampled formation configurations. In this way, the whole safe region is discretized into thousands of formation configurations.

\subsection{Formation Configuration Evaluation Functions}
To discern the optimality of the sampled formation configurations, a set of cost functions has been developed, thereby achieving an integrated optimum that takes into account the constraints of formation keeping and safety.  
\subsubsection{Goal Driven Cost}
The movement of the formation is driven by the setting goal formation configuration. Considering the distance between the current and goal formation center, the goal driven cost is designed as follows:
\begin{equation}
\label{goal function}
c_{g} = k_{g} \left | \left | \bm{c_{cur}} - \bm{c_{goal}} \right |  \right | 
\end{equation}
where $k_{g}$ is the weight parameter, $\bm{c_{cur}}$ is the formation center in the current formation configuration $\bm{f_c^{cur}}$.

\subsubsection{Scale Prefer Cost}
In addition to specifying the desired shape, the desired formation scale can also be predefined. This supplementary parameter permits differentiation between two formation configurations that display identical levels of safety with the same $\bm{c}$. Hence, the formation scale should lie within close approximation of the preferred scale:
\begin{equation}
\label{scale prefer function}
c_{s} = k_{s} \left | s_{cur}-s_{des} \right | 
\end{equation}
where $k_{s}$ is the weight parameter, $s_{cur}$ is the scale in $f_c^{cur}$ and $s_{des}$ is the desired scale.
\subsubsection{Safe Region Cost}
Once all formation configurations in one sequence are connected, the generated guidance paths based on the sequence enable the formation to maintain its desired shape without being disrupted by obstacles. To achieve this, the formation connectivity criteria (\ref{formation connectivity equ}) can be transformed into the safe region cost. For each formation configuration, the safe region cost $c_{safe}$ can be defined as follows:
\begin{equation}
\label{safe region cost}
\begin{aligned}
    c_{\text{safe}} &= \sum_{i}^{N} g_{\text{safe}}^{i}\\
    g_{\text{safe}}^{i} &=
    \begin{cases}
      k_{\text{safe}}, & \bm{p}_f^i \notin \mathcal{P}_i \\
      0, & \bm{p}_f^i \in \mathcal{P}_i
    \end{cases}
\end{aligned}
\end{equation}
where $k_{safe}$ is a large punishment cost added to the unsafe formation configuration target, and $g_{\text{safe}}^{i}$ is the self safe cost in terms of each formation target in a formation configuration.

\subsubsection{Safe Risk Cost}
During the traversal of a tightly packed environment, each UAV owns different size of free space to bypass obstacles. To render this metric more concrete, it is repurposed into a measure of risk magnitude, which can be assessed by calculating the minimal distance to the obstacles. In the context of formation maintenance, UAVs that exhibit high levels of risk are characterized by limited mobility due to the constraints imposed by their risk profiles. In contrast, UAVs with lower risk profiles are required to adjust their positions to accommodate those with higher risk levels. To quantify the risk impact, the safety risk cost, denoted as $c_{\text{risk}}$, is designed as follows:

\begin{equation}
\label{safe risk cost}
\begin{aligned}
    c_{\text{risk}} &= \sum_{i}^{N} g_{\text{risk}}^{i}\\
    g_{\text{risk}}^{i} &= w_{risk}^i\left | \left | \bm{f_t^i}-\bm{p_i}   \right |  \right | \\
    w_{risk}^i &= \begin{cases}
      \frac{d_{obs}^{i}}{\sum_{j=0}^{N} d_{obs}^{j}}, & d_{obs}^{i} >  d_{risk} \\
      0, & d_{obs}^{i} \le d_{risk}
    \end{cases}
\end{aligned}
\end{equation}
where $g_{\text{risk}}^{i}$ represents the individual safe cost as appraised for each UAV, $\bm{f_t^i}$ is the temporary calculated target of UAV$i$ and $\bm{p_i}$ is the current position of UAV$i$. The term $w_{risk}^i$ denotes the risk degree of UAV $i$ as perceived within the whole formation. Meanwhile, $d_{obs}^{i}$ is the smallest distance to the nearest obstacle from the current position, and $d_{risk}$ refers to the risk balance distance parameter. Notably, in situations where a UAV is dangerously close to an obstacle (when $d_{obs}^{i} \le d_{risk}$), the UAV's target should aim to navigate it away from the obstacle, rather than towards it. Hence, in such cases, we set the risk degree to zero.

\subsubsection{Continuity Cost}
To incorporate the continuity of the formation sequence, the proximate two formation configurations should feature comparable $c$ and $s$ values. Given that the propagation distance remains nearly constant between two formation centers, we place particular emphasis on the angle between them. Each formation configuration in the entire sequence carries a continuity cost $c_{con}$, which is derived from both the scale continuity and the angle continuity costs:
\begin{equation}
\label{continuity function}
c_{con} = k_{sc} \left | s_{cur}-s_{las} \right | +k_{ac}  \left | \theta _{cur}-\theta _{las} \right |
\end{equation}
where $k_{sc}$ and $k_{ac}$ represent the scale and angle continuity weight parameters respectively, $s_{las}$ denotes the scale of $f_c$ for the preceding formation. The current angle difference $\theta _{cur}$, can be calculated using the direction vector of the current $\bm{c}$ and the preceding $\mathbf{c}$ of $f_c$. Similarly, the preceding angle difference, $\theta _{last}$, can be calculated in the same manner.

\subsubsection{Costs Summary}
Thus far, we have discussed a number of evaluation cost functions in (\ref{goal function})-(\ref{safe risk cost}), which are pertinent to optimization in the current step, as well as (\ref{continuity function}), which strives for enhanced performance throughout the entire sequence. The total cost of a single $\bm{f_c}$ can be calculated as the sum of the above costs.

\textit{Remark 1:}
It is crucial to note that if the current sequence propagation is deemed unsafe, it is terminated to avoid potential conflict with obstacles. Nevertheless, guidance spanning a short distance is not our intended aim, as it could lead the entire formation to get trapped in a local minimum. Bearing this in mind, a substantial penalty $c_{e}$ is imposed on any halted formation sequence before the pre-defined total propagation steps, according to the remaining step number, which is a cost from a holistic perspective of the formation configuration sequence.

\subsection{Task Assignment}
The formation targets calculated from $\bm{f_c}$ could lead to potential collision among UAVs, which could be addressed by the task assignment. Despite considering temporary constant target assignment during the evaluation process, the optimal allocation cannot be guaranteed using such a method, and some intersecting paths may arise from their current positions. Consequently, we need to manage the optimal task assignment within the formation configuration to deconflict the formation. The task assignment problem is formulated as follows:
\begin{equation}
\label{task assignment}
    \min_{\sigma } \sum_{i=0}^{N} \left \| \bm{p_t^i}-\bm{p_f^{\sigma (i)}}   \right \| 
\end{equation}
where $\sigma$ is the task assignment result (eg.$\sigma(i)=j$).

Problem (\ref{task assignment}) is defined as a standard linear sum assignment problem, which can be addressed using various prevalent methods, such as the Hungarian algorithm\cite{chopra2017distributed}, simplex algorithm\cite{burger2012distributed}, auction algorithm\cite{bertsekas1989auction}. In this paper, the auction algorithm is adopted considering the computational load and a desire for optimal results.

\subsection{Algorithm Summary}
\begin{figure}[!t]
\centering
\includegraphics[width=3.4in]{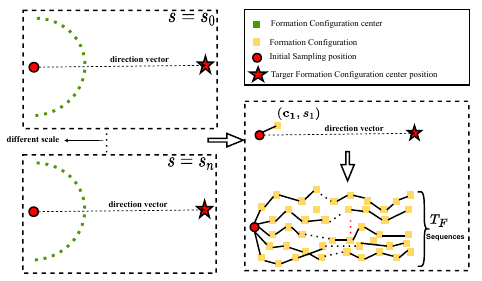}
\caption{Sampling process illustration in 2-D space.}
\label{sequence sampling illustration}
\end{figure}

Up to this point, we have presented all the modules comprising this algorithm. The complete procedure is depicted in Algorithm. \ref{SFCFG}. The algorithm takes several inputs, including the current positions of all UAVs $\mathbf{P}{cur}$, the composite formation polytope $\mathcal{P}_{all} = \cup \mathcal{P}_{i},i\in \left \{ 1, \cdots, N \right \}$, the previous optimal task assignment result $\sigma_0$, the initial sampling center $\mathbf{c_0}$, the desired target formation configuration $f_c^g$, the total number of sampling sequences $T_F$, and the propagation step $T_S$. The output of the algorithm is the generated formation path $\mathbf{\Pi }_c= \{\pi _1,\dots,\pi _N\}$ for all UAVs.
Employing the leader-follower concept, the position of the leader is assigned as $\mathbf{c_0}$, and the $s$ value in the first member of the previously generated formation configuration sequence is designated as $s_0$ respectively. The entire calculation process is centralized and executed on the leader UAV. 
Initially, $T_F$ sampled formation configuration sequences are obtained with multiple iterations of SampleFC(), which is elaborated in Section \ref{sample process}. The sequence sampling process is illustrated in Fig. \ref{sequence sampling illustration}. During the sampling process, we assess the cost of each formation configuration and get the current best step with lowest single step cost. The propagation of the current sequence is executed only when the current $\mathbf{f_{c_i}^j}$ is deemed safe (Lines 4-15). Subsequently, the suspension cost is incorporated and the total cost for each sequence is calculated (Lines 16-17). Besides, the formation configuration sequence with the lowest cost is selected as the result sequence. By solving the task assignment problem, we can obtain the optimal task assignment result based on  the first $\mathbf{f_c}$ in $\mathbf{F_c}$ and $\mathbf{P}_{cur}$. Finally, the discrete formation configurations are linked and  the result $\mathbf{\Pi }_c$ is obtained with $\sigma ^*$ and $\mathbf{F_c}$ for all UAVs (Line 19-21).

\begin{algorithm}
\caption{Sampling-based Formation Configuration Path Generation}
\label{SFCFG}
\begin{algorithmic}[1]
\Statex \textbf{Input:}$\mathbf{c_0}$, $s_0$, $\mathcal{P}_{all}$, $\sigma_0$, $\mathbf{P}_{cur}$, $\mathbf{f_c^g}$, $T_{F}, T_{S}$
\Statex \textbf{Output:}$\mathbf{\Pi }_c$
\State \textbf{Initialize}($\mathbf{c_0}$,$s_0$,$\mathcal{P}_{all}$,$\sigma_0$,$\mathbf{P}^{cur}$,$\mathbf{f_c^g}$)
\For{$i = 0$ to $T_{F}-1$}
    \State $cost_i$ $\gets$ $0$, $t_S$ $\gets$ $0$
    \For{$j = 0$ to $T_{S}-1$}
        \State $\mathbf{f_{c}^j}$ $\gets$ \textbf{SampleFC}($\mathbf{c_0}$)
        \State $cost_i^j \gets$ \textbf{FC\_Evaluation}($\mathbf{f_{c_i}^j}$)

        \If{\textbf{SafeCheck}($\mathbf{f_{c_i}^j}$)}
        \State $\mathbf{f_{c_i}^j} \gets \mathbf{f_{c}^j}$
        \State $cost_i = cost_i + cost_i^j$
        \State $\mathbf{f_c^{last}} \gets \mathbf{f_{c}^j}$
        \State $t_S ++$
        \Else
        \State \text{break}
        \EndIf
    \EndFor
    \State $cost_i = cost_i + $ \textbf{SuspendendPunish}($t_S$)
    \State $F_{c_i} \gets \{ \{f_{c_i}^0, \cdots ,f_{c_i}^{p_{step}-1}\}, cost_i    \}$
\EndFor
\State $\mathbf{F_c} \gets \textbf{SequenceGeneration}(F_{c_0}, \cdots, F_{c_{T_{F}-1}})$
\State $\sigma ^* \gets$ \textbf{TaskAssignment($\mathbf{F_c}$, $\mathbf{P}_{cur}$)}
\State $\mathbf{\Pi }_c \gets$ \textbf{PathGeneration($\mathbf{F_c}$, $\sigma ^*
$)}
\State \textbf{return} $\mathbf{\Pi }_c$
\end{algorithmic}
\end{algorithm}

\section{Distributed Formation Trajectory Optimization Based on MPPI } \label{MPPI}
The formation path generated in Section \ref{formation path} provides effective guidance for maintaining the formation and avoiding obstacles. However, these paths, denoted as $\mathbf{\Pi }_c$, consist of discrete waypoints that are connected by straight lines. Consequently, these paths may be too coarse to be directly executed in a real system. In this section, some sub-evaluation objectives of a trajectory are formulated as the components of the cost functions in (\ref{value function}). Additionally, a brief introduction to the Model Predictive Path Integral (MPPI) technique is presented. Finally, the complete trajectory optimization algorithm is summarize.
\subsection{Objective Functions}
For the sake of simplicity, the terminal cost $\phi$ is set as zero. Besides, the main components of the running cost $\rho(\bm{x}_t, t)$ presented in (\ref{value function}) are formulated as following:

\subsubsection{Formation Guidance cost}
To keep the desired formation, formation guidance cost is introduced to drive the predicted trajectory to be close to the formation path. For UAV$i$, it is formulated as follow:
\begin{equation}
h_{f}^i = \sum_{k=0}^{P-1} k_f\left | \left | \bm{p_k^i}  -\bm{p_d^{i,k}} \right |  \right | 
\end{equation}
where $k_f\in \mathbb{R}^+ $ is the positive weight and $P=T/\Delta t \in \mathbb{R}^+$ is the total propagation step number in the trajectory generation process. For better tracking performance, the $T_{S}$ in Algorithm.\ref{SFCFG} is set be equal to $P$. $\bm{p_k^i}$ is the position at step $k$ along the predicted trajectory and $\bm{p_d^{i,k}}$ is the $k_{th}$ formation guidance waypoint extracted from the received guidance path of UAV$i$.
\subsubsection{Dynamic Cost}
The trajectory is limited by the dynamics of UAV in terms of the maximum velocity and acceleration. For UAV$i$ it is defined as:
\begin{equation}
\begin{aligned}
    h_{\text{d}}^{i} &=
    \begin{cases}
      k_{\text{dyn}}, & d_{dyn}^i\ge  0 \\
      0, & d_{dyn}^i<  0
    \end{cases}\\
d_{dyn}^i&=sign(\left | v^i_t \right | -v_{max} )+sign(\left | a^i_t \right | -a_{max} )
\end{aligned}
\end{equation}
where $v^i_t$ and $a^i_t$ are the velocity and acceleration at time $t$ along the trajectory, $v_{max}$ and $a_{max}$ are the corresponding maximum and $k_{\text{dyn}}\in \mathbb{R}^+ $.
\subsubsection{Smoothness Cost}
The smooth trajectory is required in the flight, therefore the smoothness cost is designed to penalize the integral square derivatives of control input $\bm{u}_t^i$. It is designed as follows:
\begin{equation}
h_S=k_{smo}\int_{t}^{t+T}{\bm{u}_\tau ^i}^{\text{T}}\bm{u}_\tau ^id\tau 
\end{equation}
where $k_{smo} \in \mathbb{R}^+ $ is the smoothness weight.
\subsubsection{Obstacle Avoidance Cost}
 To ensure safety, obstacle avoidance cost is incorporated to keep UAV far away from the obstacles. For UAV$i$, it is expressed as:
 \begin{equation}
h_{safe}^{i}=\left\{ \begin{matrix}
   k_{obs}, & d_{obs}^{i}<d_{obs }^{min}  \\
   k_{obs}{{\left( \frac{d_{obs }^{max} -d_{obs}^i}{d_{obs }^{max}-d_{obs }^{min}} \right)}^{\beta 
 }}, & d_{obs }^{min}<  d_{obs}^{i}\le d_{obs }^{max}  \\
   0, & d_{obs}^{i}>d_{obs }^{max}  \\
\end{matrix} \right.
\end{equation}
where $d_{obs }^{min}$ and $d_{obs }^{max}$ are the safe threshold of obstacles avoidance,$k_{obs},\beta>0 \in \mathbb{R}^+$.
\subsubsection{Mutual Collision Cost}
It is essential to avoid the inner collion during the formation flight. The mutual collision cost is designed to exert a repulsive force between two UAVs within a mutual distance less than $d_{mut}^{max}$. For UAV $i$ and $j,j\ne i$, it is presented by:
\begin{equation}
\begin{aligned}
    h_{mut}^{i} =
    \begin{cases}
      k_{mut}, & d_{mut}^{ij}<d_{mut}^{min} \\
      k_{mut}{{\left( \frac{d_{mut }^{max} -d_{mut}^{ij}}{d_{mut}^{max}-d_{mut}^{min}} \right)}^{\alpha 
 }}, & d_{mut}^{min}<  d_{mut}^{ij}\le d_{mut}^{max} \\
0, & d_{mut}^{ij}>d_{mut}^{max}
    \end{cases}\\
d_{mut}^{ij}=\left \| \Gamma [p_i(t)-p_j(t)] \right \| _2,p_i(t)\in \Omega _i,p_j(t)\in \kappa _j
\end{aligned}
\end{equation}
where $k_{mut},\alpha>0$, $p_i(t)$ and $p_j(t)$ are the position of UAV $i,j$ at time $t$ from the predicted trajectory $\Omega _i$ and the executed trajectory $\kappa _j$ respectively.Given the downwash peril, a transformation matrix $\Gamma = \text{diag}(1,1,\lambda),\lambda<1$ is introduced to transform the euclidean distance to ellipsoidal distance between UAV $i$ and $j$ at time $t$.
\subsection{Model Predictive Path Integral Control}
To make the paper self-contained, in this part, we will give a brief introduction of MPPI, more details can be found in our previous work \cite{lu2021flight}.
The stochastic control problem presented in (\ref{stochastic optimal control problem}) can be solved with the Hamilton-Jacobo-Bellman(HJB) equation. Thus the HJB equation based on the dynamics model(\ref{dynamics model}) and cost functions(\ref{value function}) is defined as follows:
\begin{equation}
\label{PDE}
\begin{split}
- \frac{{\partial V_i}}{{\partial t}} & = \rho (\bm{x}_t^i,t) + {\left( {A \bm{x}_t^i} \right)^T}\frac{{\partial V_i}}{{\partial {\bm{x}_t^i}}} 
 - \frac{1}{2}{\left( {\frac{{\partial V_i}}{{\partial {\bm{x}_t^i}}}} \right)^T}B{R^{ - 1}}{B^T} {\frac{{\partial V_i}}{{\partial {\bm{x}_t^i}}}}  \\
& + \frac{1}{2}{\rm{tr}}\left( {B{B^T}\frac{{{\partial ^2}V}}{{\partial {\bm{x}_t^i}^2}}} \right)
\end{split}
\end{equation}
with the boundary condition $V_i(\bm{x}_{t+T}^i,t+T)=\phi (\bm{x}_{t+T},t+T)=0$. The optimal control input is expressed as follows:
\begin{equation}
{\bm{u}_t^i} =  - {R^{ - 1}}{B^T}\frac{{\partial V_i}}{{\partial {\bm{x}_t^i}}}
\end{equation}

By solving the partial differential equation (PDE) of cost function $V_i(\bm{x}_t^i,t)$ to state vector $\bm{x}_t^i$, $\bm{u_t^i}$ is obtained. However, it is nontrivial to solve such a problem with high dimension of $\bm{x}_t^i$ with tradition methods. Inspired by the path integral control framework, the PDE can be transformed into a path integral. For the sake of utilization of the Fetnman-Kac formula relating PDE to path integrals, (\ref{PDE}) is needed to by linear. Thus (\ref{PDE}) is linearized with  exponential transform:
\begin{equation}
\begin{array}{*{20}{c}}
\frac{{\partial \Psi_i }}{{\partial t}} = \frac{\Psi_i }{\lambda }\rho (\bm{x}_t^i,t) - {\left( {A \bm{x}_t^i} \right)^T}\frac{{\partial \Psi_i }}{{\partial {\bm{x}_t^i}}} - \frac{1}{2}tr\left( {B{B^T}\frac{{{\partial ^2}\Psi_i }}{{\partial {\bm{x}_t^i}^2}}} \right)\\
\Psi_i \left( {{\bm{x}_t^i},t} \right) = {e^{ - \frac{{V_i(\bm{x}_t^i,t)}}{\lambda }}}
\end{array}
\end{equation}
where $\lambda \in \mathbb{R^+} $. Then the Fetnman-Kac formula is applied to (\ref{PDE}) denoted by:
\begin{equation}
\label{FK}
\begin{aligned} 
\Psi_i \left( {{\bm{x}_t^i},t} \right)=\mathbb{E} [exp(-\frac{1}{\lambda } )]S(t)\\
S(t)=\int_{t}^{t+T}\rho (\bm{x_\tau ^i},\tau )dt 
\end{aligned}
\end{equation}

Now the path integral form for the cost function in (\ref{FK}) is obtained through the above process. By computing the gradient of $\Psi_i \left( {{\bm{x}_t^i},t}\right)$ with respect to the initial state $\bm{x}_{t}^i$ analytically, we can get follows:
\begin{equation}
{{{\boldsymbol u}_t^i}^*}dt = {R^{ - 1}}{B^T}{\left( {B{R^{ - 1}}{B^T}} \right)^{ - 1}}\frac{{\mathbb {E}\left[ {\exp \left( { - \frac{{S\left( t \right)B}}{\lambda }d{\boldsymbol v}_t^i} \right)} \right]}}{{\mathbb {E}\left[ {\exp \left( { - \frac{{S\left( t \right)}}{\lambda }} \right)} \right]}}
\end{equation}

Subsequently, the dynamic model for trajectory sampling is discretized as follows:
\begin{equation}
    d\bm{x_{t+1}} =A\bm{x_t}\Delta t+B\mathbf{\mu _t}\Delta t
\end{equation}

Finally, the discretized expression of ${{{\boldsymbol u}_t^i}^*}$ is derived:
\begin{equation}
\begin{aligned} 
  {{{\boldsymbol u}_t^i}^*}={{{\boldsymbol u}_{t-1}^i}}+\frac{{\mathbb {E}\left[ {\exp \left( { - \frac{{S\left( t+T \right)}}{\lambda }d{\boldsymbol v}_t^i} \right)} \right]}}{{\mathbb {E}\left[ {\exp \left( { - \frac{{S\left( t+T \right)}}{\lambda }} \right)} \right]}}\\
\end{aligned}
\end{equation}
where $S(t+T)=\sum_{m=0}^{R} \rho (\bm{x}_{m,t}^i,t)\Delta t$, $\Delta t$ is the time interval and $R$ is the total number of sampled trajectories.

\subsection{Distributed Formation Trajectory Optimization}
As delineated through the aforementioned predictive inference control process, it is feasible to ascertain the optimal control input. Consequently, this leads to the extraction of the optimal trajectory, which is fundamentally based on the state vector \(\bm{x_t}\) at the time \(t\) and the optimal control sequence \(U_i^*\).

For the UAV$i$, the trajectory optimization approach, as proposed in this study, is succinctly summarized in Algorithm. \ref{MPPI alg}. The algorithm comprises several variables: the current state vector \(\bm{x} ^i_{t}\), initial control input \(\bm{u}^i_{init}\), initial control input sequence \(U_i^{init}\), total propagation steps \(P\), \(R\) and the alternative executed trajectories set signified as \(\mathbb{K}=\{\kappa _j,\dots,\kappa _N\},j\ne i\). The output of the algorithm is the optimally generated trajectory $\kappa_i$ by UAV$i$.
The algorithm initiates by setting the initial state vector \(\bm{x} ^i_{0}\) equal to \(\bm{x} ^i_{t}\). \(R\) distinct trajectories are generated by employing the random sampling set \(\mathbb{V}^i\) where the \(m\)th trajectory is explicitly denoted as \(\mathbb{V}^i_m\). These trajectories are generated parallelly over \(P\) steps integrating both the control input \(\bm{u}_{t}^i\) and disturbance \(\bm{v_{t}^i}\). Simultaneously, the trajectories undergo assessment in the \(m\)th Monte Carlo sampling process by the function \(S(\mathbb{V}_m^i)\) (Line 2-8).

Subsequently, the importance weight \(\omega(\mathbb{V}_m^i)\) is computed for each trajectory based on the expectation cost (Line 9-13), leading to the acquisition of optimal control inputs (Line 9-17). The states of the optimal trajectory are then obtained by incorporating \(u^*\) within the dynamic model. Moreover, the process involves the iterative update of the optimal control sequence, where the current \(U_i^*\) is employed as the initial control sequence in the succeeding calculation round (Line 20-24).
To amplify the practicality of the method, the obtained trajectory states undergo the processes of connection and smoothing operations, thereby yielding the final optimal trajectory as the output (Line 25). 
\begin{algorithm}
\caption{ Distributed Multi-UAV Trajectory optimization based on MPPI}
\label{MPPI alg}
\begin{algorithmic}[1]
\Statex \textbf{Input:}$\bm{x} ^i_{t}$,$\bm{u}^i_{init}$, $U_i^{init}$, $P$, $R$, $\mathbb{K}$
\Statex \textbf{Output:}$\kappa_i$
\State $ x_{0}^i = \bm{x} ^i_{t}$
\For{$m = 0$ to $R-1$}
    \State $\mathbb{V}_m^i \gets$\{${\bm{v_{m,0}^i},\bm{v_{m,1}^i},\dots,\bm{v_{m,P-1}^i}}$\}
    \For{$n = 0$ to $P-1$}
        \State $\mathbf{ x_{n+1}^i} = A\bm{x_{n}^i} + B(\bm{u}_{t}^i+\bm{v_{t}^i})$ 
        \State $S(\mathbb{V}_m^i) += S(\mathbb{V}_m^i) + \rho (\bm{x_{n+1} ^i},t_n)$
    \EndFor
\EndFor
\State $S_{min} \gets \min_{}{\{S(\mathbb{V}_0^i),S(\mathbb{V}_1^i),\dots S(\mathbb{V}_{R-1}^i)\}}$
\State $\varepsilon \gets \sum_{m=0}^{R} exp(-\frac{1}{\lambda }(S(\mathbb{V}_m^i)-S_{min} ))$
\For{$m = 0$ to $R-1$}
\State $ \omega(\mathbb{V}_m^i)\gets \frac{1}{\varepsilon }  exp(-\frac{1}{\lambda }(S(\mathbb{V}_m^i)-S_{min} ))$
\EndFor
\For{$n = 0$ to $P-1$}
\State ${\bm{u_n^i}}^* = \bm{u_n^i} +\sum_{m=0}^{R-1} w(\mathbb{V}_m^i) \bm{v_{m,n}^i}$
\EndFor
\For{$n = 0$ to $P-1$}
\State ${\bm{x_{n+1}^i}}^* = A{\bm{x_{n}^i}}^* + B{\bm{u}_{t}^i}^*$
\EndFor
\For{$n = 0$ to $P-2$}
\State ${\bm{u_n^i}}^* = {\bm{u_{n+1}^i}}^*$
\EndFor
\State ${\bm{u_{P-1}^i}}^* = \bm{u}^i_{init}$
\State $U_i^*=\{{\bm{u_0^i}}^*,\dots,{\bm{u_{P-1}^i}}^*\}$
\State return $\kappa_i \gets \textbf{Link}({\bm{x_{1}^i}}^*,{\bm{ x_{2}^i}}^*,\dots,{\bm{x_{P}^i}}^*)$

\end{algorithmic}
\end{algorithm}

\section{Result Analysis}\label{validation section}
In this section, the proposed strategy are validated in some simulation scenarios. Firstly, a short introduction of the assessments indicators for evaluating the performance of formation flight is presented. Then the criteria will be used in the later comparisons. Finally, some simulation comparisons between the developed method and the methods in \cite{zhou2018agile}, \cite{quan2022distributed} are conducted. 
\subsection{Evaluation Criteria}
The goal of the proposed algorithm is to keep the whole formation safe and fly in the desired shape. Therefore,  the formation performance is evaluated from the perspective of safety and formation similarity. More details of formation similarity can be seen in \cite{quan2022distributed}. To evaluate the whole process of formation flight, the average formation similarity error $\overline{e_{f}}$ and the max formation similarity error $e_f^{max}$ which describes the comprehensive formation keeping and the degree of formation distortion are defined as follows:
\begin{equation}
\begin{aligned}
    \overline{e_{f}} &= \frac{\int_{t_0}^{t_{e}}f(t)dt}{t_{e}-t_0}\\
e_f^{max}&=maxf(t),t\in [t_0,t_{end}]
\end{aligned}
\end{equation}
where the definition of formation similarity error $f$ can be found in \cite{quan2022distributed}, $f(t)$ is the formation similarity error at time $t$, and $t_0,t_{e}$ are the formation start time and end time during the whole flight process. The success rate $suc (\%)$ is also utilized to evaluate the degree of safety and formation keeping. Once the formation collides with the obstacles or distorts seriously, it is failed.
\subsection{Simulation Experiment}

\begin{figure*}[!t]
\centering
\includegraphics[width=7.0in]{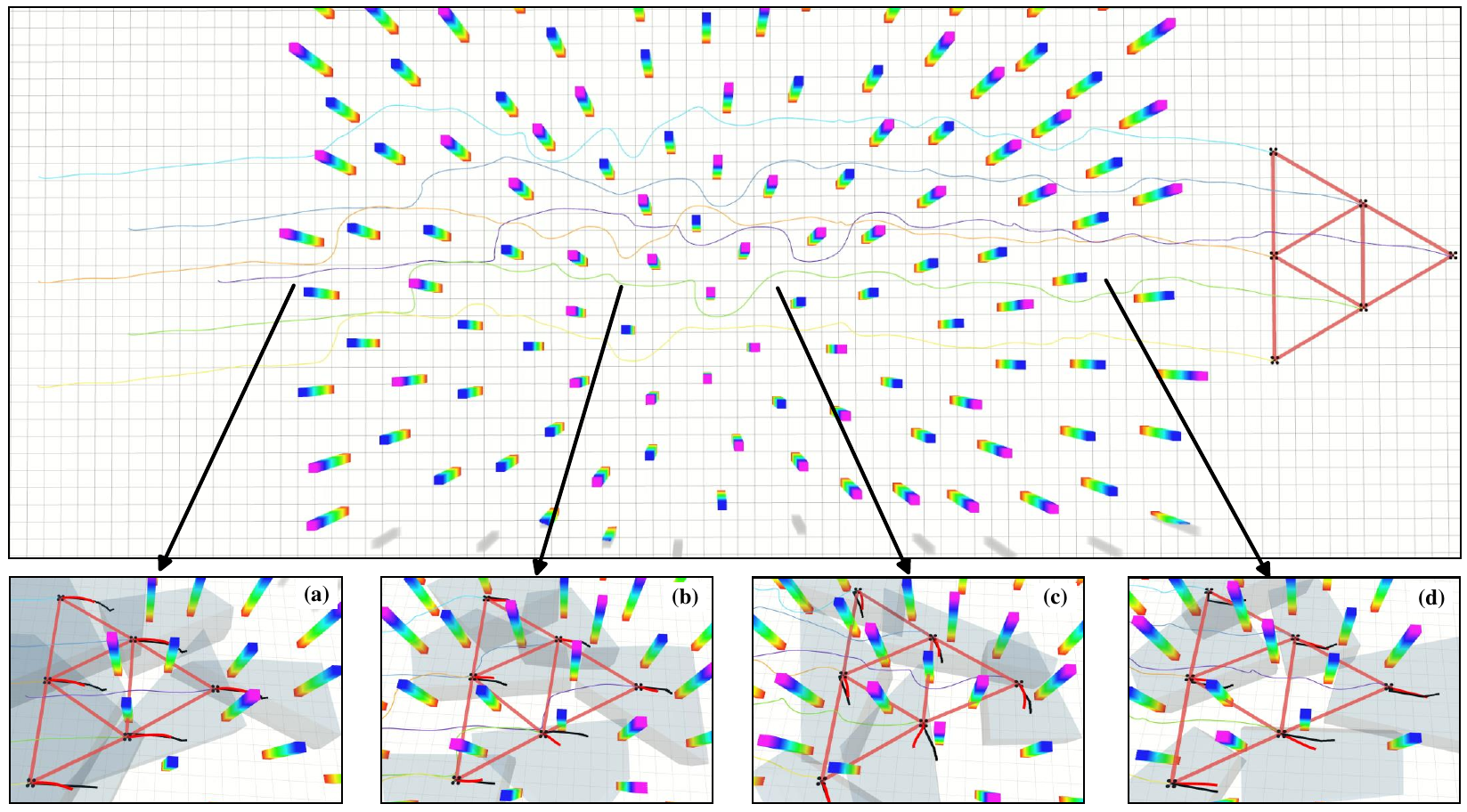}
\caption{A triangular formation, consisting of six unmanned aerial vehicles (UAVs), traverses through an unknown environment with obstacles from the left side to the right side. (a)-(d) are the snapshots in the fixed time interval.}
\label{all_process_with_snap_2}
\end{figure*}

\begin{figure*}[!t]
\centering

\includegraphics[width=7.0in]{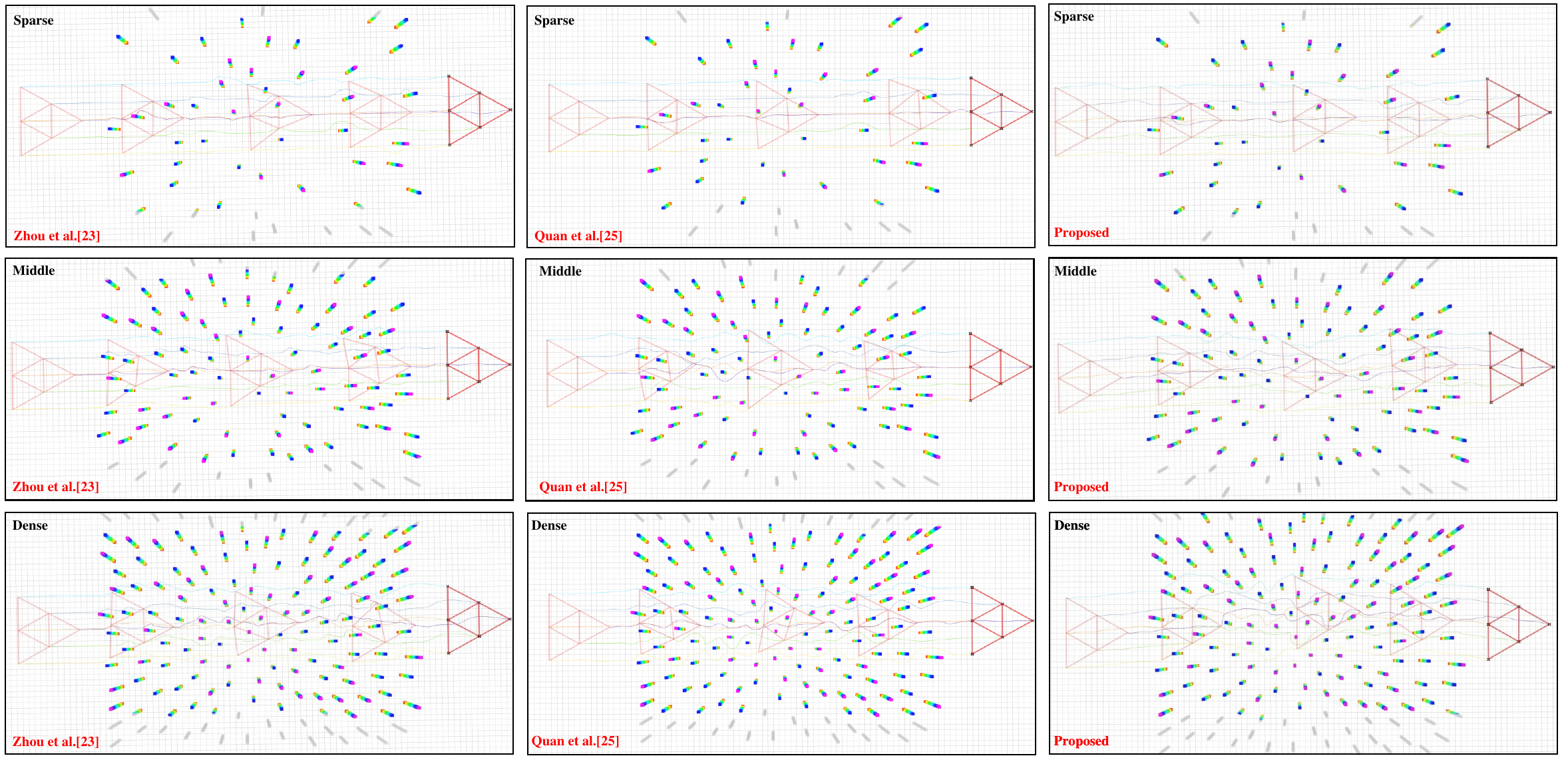}
\caption{The formation flight process generated by our proposed method, \cite{zhou2018agile} and \cite{quan2022distributed} in environments with different number of obstacles.}
\label{cluster result}
\end{figure*}

The simulations are conducted in cluster environments measuring $50 \times 40$ m, with varying numbers of pillar obstacles (50, 100, 150). Additionally, a narrow corridor environment measuring $26 \times 8$ m is created to assess the scale adaptability of the formation keeping methods. The narrow width of the corridor presents a significant challenge for scale adjustment. The maximum velocity is set to $1.5$ m/s, and a triangle formation with 6 UAVs is the desired formation. All simulations are performed on a computer equipped with an Intel i7-12700F CPU and NVIDIA GeForce RTX 3070 GPU. 

\begin{figure}[!t]
\centering
\subfloat[]{\includegraphics[width=3.4in]{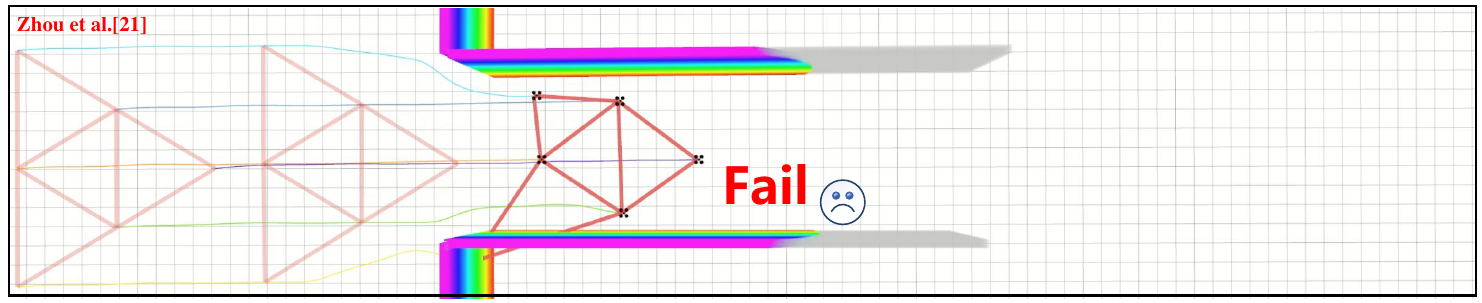}\label{vrb fail in narrow corridor}
}
\hfil
\subfloat[]{\includegraphics[width=3.4in]{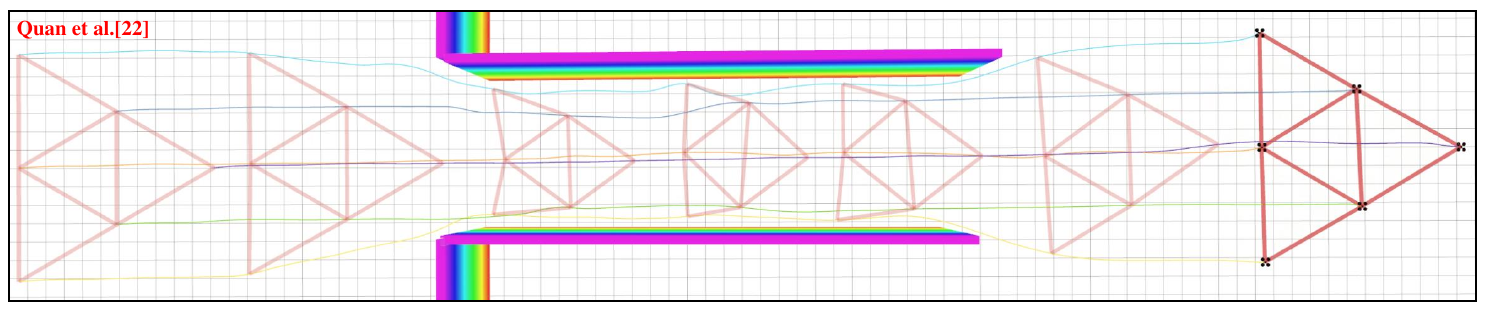}\label{quan in narrow corridor}
}
\hfil
\subfloat[]{\includegraphics[width=3.4in]{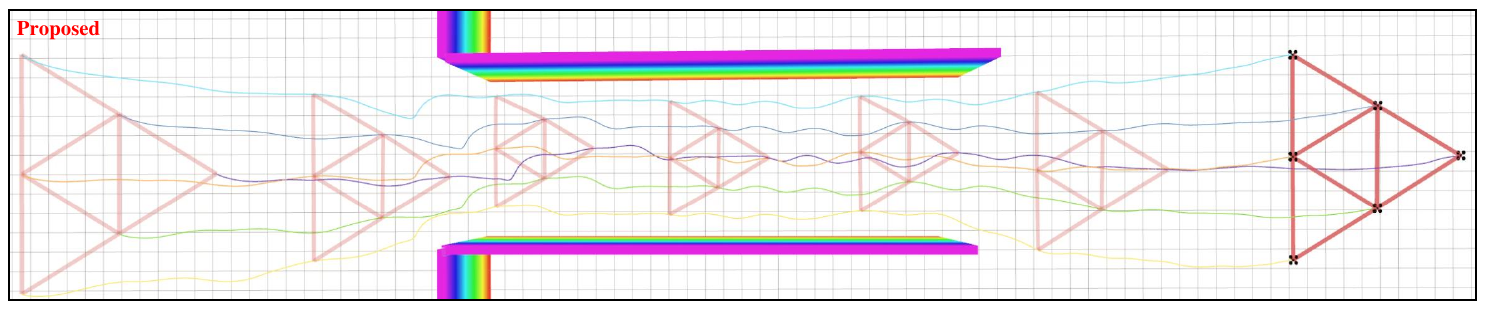}\label{my method in narrow corridor}
}
\caption{The shallow red line are the formation trace left every fixed distance. Fig \ref{vrb fail in narrow corridor}, Fig \ref{quan in narrow corridor}  and Fig \ref{my method in narrow corridor} are the the flight process with the method in \cite{zhou2018agile}, \cite{quan2022distributed} and the proposed method respectively.
}
\label{narrow corridor result}
\end{figure}

As illustrated in Fig. \ref{all_process_with_snap_2}, the formation employing our proposed method successfully navigates through an unknown environment containing obstacles.
The comparison results of the simulations in the cluster environments and the narrow corridor are depicted in Fig. \ref{cluster result} and Fig. \ref{narrow corridor result} respectively. The flight results are summarized in Table \ref{simulation result}, and it can be observed that all methods, including the proposed method and the ones presented in \cite{zhou2018agile} and \cite{quan2022distributed}, successfully navigate through the environments while maintaining the formation. The formation keeping is presented as a formation vector field in \cite{zhou2018agile}  and one aspect of gradient optimization problem in \cite{quan2022distributed} respectively, which is optimized concurrently with other aspects, such as collision evasion, obstacles avoidance and trajectory smoothness. Consequently, the influence of formation constraints can be diminished to prioritize individual security. The efficacy of formation maintenance is observed to diminish as the number of obstacles escalates in the methodologies presented by \cite{zhou2018agile} and \cite{quan2022distributed}, as evidenced by the significant increase of $\overline{e_{f}}$ and $e_f^{max}$ in different clustered scenarios, which is displayed in Table \ref{simulation result}. In the proposed method, the issue of weakened formation preservation is circumvented, as the sampled formation configurations are confined to the local safe region, thereby ensuring formation connectivity and obstacle avoidance as well.
\begin{table}[h]
\caption{The performance comparison in different environments}
\centering
\begin{tabular}{|c|c|c|c|c|}
\hline
\textbf{Environment} & \textbf{Method} & $suc (\%)$ & $\overline{e_{f}}$ & $e_f^{max}$\\ \hline
\multirow{3}{*}{Sparse} & Zhou et al.\cite{zhou2018agile} & 100 & 0.0025 & 0.0268 \\  
   & Quan et al.\cite{quan2022distributed} & 100 & 0.0013 & 0.0109 \\  
   & Proposed & 100 & \textbf{0.0011}& \textbf{0.0057}  \\  \hline 
\multirow{3}{*}{Middle} & Zhou et al.\cite{zhou2018agile} & 100 & 0.0036 & 0.0315  \\  
   & Quan et al.\cite{quan2022distributed} & 100 & 0.0052 & 0.0327 \\  
   & Proposed & 100 & \textbf{0.0012}& \textbf{0.0064} \\  \hline
\multirow{3}{*}{Dense} & Zhou et al.\cite{zhou2018agile} & 100 & 0.0059 & 0.0586  \\  
   & Quan et al.\cite{quan2022distributed} & 100 & 0.0094 & 0.0456 \\  
   & Proposed & 100 & \textbf{0.0020}& \textbf{0.0071} \\  \hline
\multirow{3}{*}{Narrow Corridor} & Zhou et al.\cite{zhou2018agile} & 0 & - & -  \\  
   & Quan et al.\cite{quan2022distributed} & 100 & 0.0102 & 0.0492  \\  
   & Proposed & 100 & \textbf{0.0007}& \textbf{0.0134} \\  \hline
\end{tabular}
\label{simulation result}
\end{table}

Moreover, the scalability performance of the method in \cite{zhou2018agile} is inadequate, particularly in the narrow corridor environment, where the formation fails to adjust its scale freely. This limitation leads to unsuccessful flights in such challenging conditions. Conversely, when the safe region within the confined corridor is too restricted, the scale of the formation is adeptly adjusted to as low as 0.5, thereby exemplifying the superior adaptability of the formation in challenging environments, as achieved by the proposed method. Furthermore,   $\overline{e_{f}}$ and $e_f^{max}$ achieved with the proposed method is significantly reduced compared to the methods in \cite{zhou2018agile} and \cite{quan2022distributed} in both clustered and narrow corridor environments, which suggests the proposed method effectively mitigates formation distortion.
In conclusion, the proposed method behave better in environments with various obstacles.




\section{Conclusion}\label{conclusion}
In this paper, a sampling-based hierarchical trajectory planning method is proposed, consisting of front-end and back-end approaches, to achieve safe UAV formation flight with the desired shape. The formation safe region can be defined by transmitting the safe flight corridor, which represents the limited safe area with a small data volume. Subsequently, the formation guidance paths are derived through a sampling-based front-end path generation method, wherein  numerous formation configurations stochastically sampled. The optimal sequence of formation configurations is then identified through the application of specific cost functions, from which the guidance paths are generated. Furthermore, a distributed trajectory optimization technique based on MPPI is introduced as the back-end method, which guarantees the generation of smooth and safe trajectories. Owing to the parallel computational capabilities afforded by the GPU, the proposed sampling-based method is capable of functioning in real-time. Finally, extensive simulations are conducted to validate the proposed framework, and comparisons are made with two other state of the art algorithms. The result demonstrates that the proposed method enhances the overall performance of the formation in environments with challenging obstacles. In the future, we aim to further develop the proposed method into a fully distributed approach and apply it to real-world scenarios, such as goods delivery in unknown environments.

\bibliographystyle{IEEEtran}
\bibliography{arxiv}

\end{document}